# On the Robustness of Most Probable Explanations

**Hei Chan***
School of Electrical Engineering and Computer Science
Oregon State University
Corvallis, OR 97330
chanhe@eecs.oregonstate.edu

**Adnan Darwiche**
Computer Science Department
University of California, Los Angeles
Los Angeles, CA 90095
darwiche@cs.ucla.edu

## Abstract

In Bayesian networks, a Most Probable Explanation (MPE) is a complete variable instantiation with the highest probability given the current evidence. In this paper, we discuss the problem of finding robustness conditions of the MPE under single parameter changes. Specifically, we ask the question: How much change in a single network parameter can we afford to apply while keeping the MPE unchanged? We will describe a procedure, which is the first of its kind, that computes this answer for all parameters in the Bayesian network in time $O(n \exp(w))$, where $n$ is the number of network variables and $w$ is its treewidth.

## 1 Introduction

A *Most Probable Explanation (MPE)* in a Bayesian network is a complete variable instantiation which has the highest probability given current evidence [1]. Given an MPE solution for some piece of evidence, we concern ourselves in this paper with the following question: What is the amount of change one can apply to some network parameter without changing this current MPE solution? Our goal is then to deduce robustness conditions for MPE under single parameter changes. This problem falls into the realm of *sensitivity analysis*. Here, we treat the Bayesian network as a system which accepts network parameters as inputs, and produces the MPE as an output. Our goal is then to characterize conditions under which the output is guaranteed to be the same (or different) given a change in some input value.

This question is very useful in a number of application areas, including what-if analysis, in addition to the design and debugging of Bayesian networks. For an example, consider Figure 1 which depicts a Bayesian network for diagnosing potential problems in a car. Suppose now that we have the following evidence: the dashboard test and the lights test came out positive, while the engine test came out negative. When we compute the MPE in this case, we get a scenario in which all car components are working normally. This seems to be counterintuitive as we expect the most likely scenario to indicate at least that the engine is not working. The methods developed in this paper can be used to debug this scenario. In particular, we will be able to identify the amount of change in each network parameter which is necessary to produce a different MPE solution. We will revisit this example later in the paper and discuss the specific recommendations computed by our proposed algorithm.

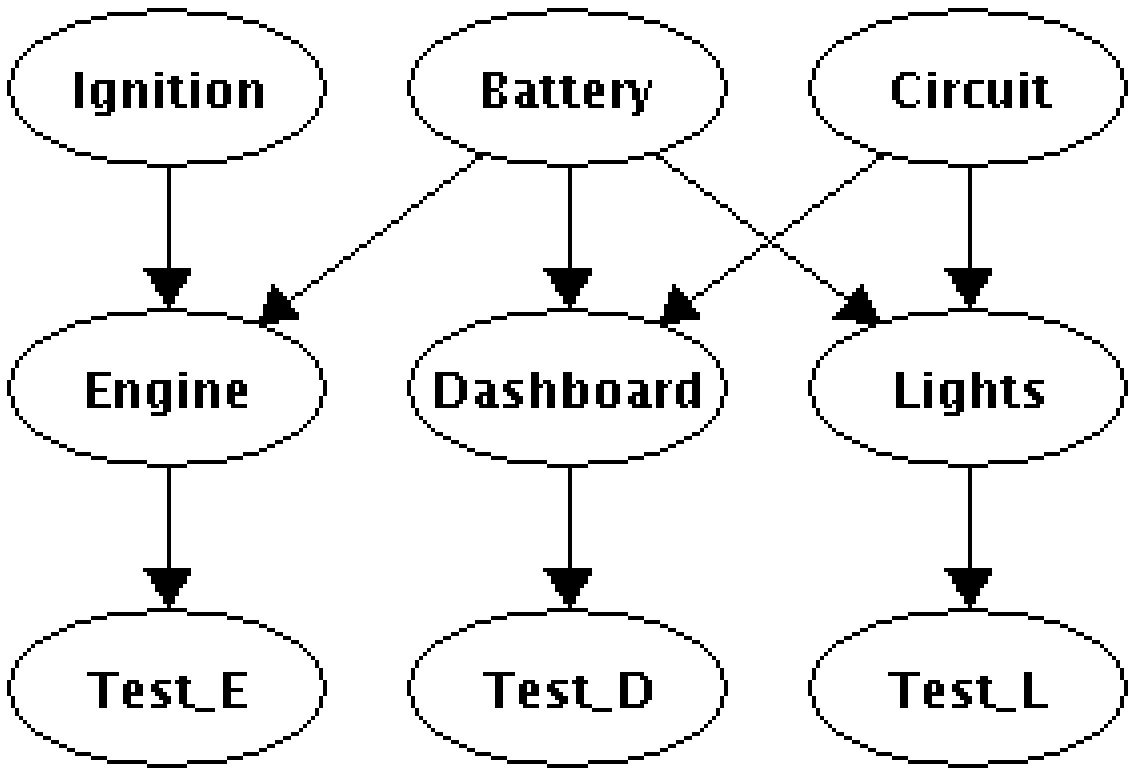

Figure 1: An example Bayesian network where we are interested in the MPE and its robustness.

Previous results on sensitivity analysis have focused mostly on the robustness of probability values, such as the probability of evidence, under single or multiple parameter changes [2, 3, 4, 5, 6, 7, 8, 9]. Because probability values are continuous, while MPE solutions are discrete instantiations, abrupt changes in MPE so-

*This work was completed while Hei Chan was at UCLA.

lutions may occur when we change a parameter value. This makes the sensitivity analysis of MPE quite different from previous work on the subject.

This paper is structured as follows. We first provide the formal definition of Bayesian networks and MPE in Section 2. Then in Section 3, we explore the relationship between the MPE and a single network parameter, and also look into the case where we change co-varying parameters in Section 4. We deduce that the relationship can be captured by two constants that are independent of the given parameter. Next in Section 5, we show how we can compute these constants for all network parameters, allowing us to automatically identify robustness conditions for MPE, and provide a complexity analysis of our proposed approach. Finally, we show some concrete examples in Section 6, and then extend our analysis to evidence change in Section 7.

## 2 Most Probable Explanations

We will formally define most probable explanations in this section, but we specify some of our notational conventions first. We will denote variables by uppercase letters ($X$) and their values by lowercase letters ($x$). Sets of variables will be denoted by bold-face uppercase letters ($\mathbf{X}$) and their instantiations by bold-face lowercase letters ($\mathbf{x}$). For variable $X$ and value $x$, we will often write $x$ instead of $X = x$, and hence, $Pr(x)$ instead of $Pr(X = x)$. For a binary variable $X$ with values *true* and *false*, we will use $x$ to denote $X = true$ and $\bar{x}$ to denote $X = false$. Therefore, $Pr(X = true)$ and $Pr(x)$ represent the same probability in this case. Similarly, $Pr(X = false)$ and $Pr(\bar{x})$ represent the same probability. Finally, for instantiation $\mathbf{x}$ of variables $\mathbf{X}$, we will write $\neg\mathbf{x}$ to mean the set of all instantiations $\mathbf{x}^\star \neq \mathbf{x}$ of variables $\mathbf{X}$. For example, we will write $Pr(\mathbf{x}) + Pr(\neg\mathbf{x}) = 1$.

A Bayesian network is specified by its structure, a *directed acyclic graph (DAG)*, and a set of *conditional probability tables (CPTs)*, with one CPT for each network variable [1]. In the CPT for variable $X$ with parents $\mathbf{U}$, we define a network parameter $\theta_{x|\mathbf{u}}$ for every family instantiation $x\mathbf{u}$ such that $\theta_{x|\mathbf{u}} = Pr(x \mid \mathbf{u})$.

Given the network parameters, we can compute the probability of a complete variable instantiation $\mathbf{x}$ as follows:

$$Pr(\mathbf{x}) = \prod_{x\mathbf{u} \sim \mathbf{x}} \theta_{x|\mathbf{u}}, \tag{1}$$

where $\sim$ is the compatibility relation between instantiations, i.e., $x\mathbf{u} \sim \mathbf{x}$ means that $x\mathbf{u}$ is compatible with $\mathbf{x}$). Now assume that we are given evidence $\mathbf{e}$. A *most probable explanation (MPE)* given $\mathbf{e}$ is a complete variable instantiation that is consistent with $\mathbf{e}$ and has the highest probability [1]:

$$\begin{aligned} MPE(\mathbf{e}) &\stackrel{def}{=} \arg\max_{\mathbf{x} \sim \mathbf{e}} Pr(\mathbf{x}) \\ &= \arg\max_{\mathbf{x} \sim \mathbf{e}} \prod_{x\mathbf{u} \sim \mathbf{x}} \theta_{x|\mathbf{u}}. \end{aligned} \tag{2}$$

We note that the MPE may not be a unique instantiation as there can be multiple instantiations with the same highest probability. Therefore, we will define $MPE(\mathbf{e})$ as a set of instantiations instead of just one instantiation. Moreover, we will sometimes use $MPE(\mathbf{e}, \neg\mathbf{x})$ to denote the MPE instantiations that are consistent with $\mathbf{e}$ but inconsistent with $\mathbf{x}$.

In the following discussion, we will find it necessary to distinguish between the MPE identity and the MPE probability. By the *MPE identity*, we mean the set of instantiations having the highest probability. By the *MPE probability*, we mean the probability assumed by a most likely instantiation, which is denoted by:

$$MPE_p(\mathbf{e}) \stackrel{def}{=} \max_{\mathbf{x} \sim \mathbf{e}} Pr(\mathbf{x}). \tag{3}$$

This distinction is important when discussing robustness conditions for MPE since a change in some network parameter may change the MPE probability, but not the MPE identity.

## 3 Relation Between MPE and Network Parameters

Assume that we are given evidence $\mathbf{e}$ and are able to find its MPE, $MPE(\mathbf{e})$. We now address the following question: How much change can we apply to a network parameter $\theta_{x|\mathbf{u}}$ without changing the MPE identity of evidence $\mathbf{e}$? To simplify the discussion, we will first assume that we can change this parameter without changing any co-varying parameters, such as $\theta_{\bar{x}|\mathbf{u}}$, but we will relax this assumption later.

Our solution to this problem is based on some basic observations which we discuss next. In particular, we observe that complete variable instantiations $\mathbf{x}$ which are consistent with $\mathbf{e}$ can be divided into two categories:

- Those that are consistent with $x\mathbf{u}$. From Equation 1, the probability of each such instantiation $\mathbf{x}$ is a linear function of the parameter $\theta_{x|\mathbf{u}}$.
- Those that are inconsistent with $x\mathbf{u}$. From Equation 1, the probability of each such instantiation $\mathbf{x}$ is a constant which is independent of the parameter $\theta_{x|\mathbf{u}}$.

Let us denote the first set of instantiations by $\Sigma_{\mathbf{e},x\mathbf{u}}$ and the second set by $\Sigma_{\mathbf{e},\neg(x\mathbf{u})}$. We can then conclude that:

- The set of most likely instantiations in $\Sigma_{\mathbf{e},x\mathbf{u}}$ remains unchanged regardless of the value of parameter $\theta_{x|\mathbf{u}}$, even though the probability of such instantiations may change according to the value of $\theta_{x|\mathbf{u}}$. This is because the probability of each instantiation $\mathbf{x} \in \Sigma_{\mathbf{e},x\mathbf{u}}$ is a linear function of the value of $\theta_{x|\mathbf{u}}$: $Pr(\mathbf{x}) = r \cdot \theta_{x|\mathbf{u}}$, where $r$ is a coefficient independent of the value of $\theta_{x|\mathbf{u}}$. Therefore, the relative probabilities among instantiations in $\Sigma_{\mathbf{e},x\mathbf{u}}$ remain unchanged as we change the value of $\theta_{x|\mathbf{u}}$. Note also that the most likely instantiations in this set $\Sigma_{\mathbf{e},x\mathbf{u}}$ are just $MPE(\mathbf{e}, x\mathbf{u})$ and their probability is $MPE_p(\mathbf{e}, x\mathbf{u})$. Therefore, if we define:

$$r(\mathbf{e}, x\mathbf{u}) \stackrel{def}{=} \frac{\partial MPE_p(\mathbf{e}, x\mathbf{u})}{\partial \theta_{x|\mathbf{u}}}, \tag{4}$$

  we will then have:

$$Pr(\mathbf{x}) = r(\mathbf{e}, x\mathbf{u}) \cdot \theta_{x|\mathbf{u}},$$

  for any $\mathbf{x} \in MPE(\mathbf{e}, x\mathbf{u})$.

- Both the identity and probability of the most likely instantiations in $\Sigma_{\mathbf{e},\neg(x\mathbf{u})}$ are independent of the value of parameter $\theta_{x|\mathbf{u}}$. This is because the probability of each instantiation $\mathbf{x} \in \Sigma_{\mathbf{e},\neg(x\mathbf{u})}$ is independent of the value of $\theta_{x|\mathbf{u}}$. Note that the most likely instantiation in this set $\Sigma_{\mathbf{e},\neg(x\mathbf{u})}$ is just $MPE(\mathbf{e}, \neg(x\mathbf{u}))$. We will define the probability of such an instantiation as:

$$k(\mathbf{e}, x\mathbf{u}) \stackrel{def}{=} MPE_p(\mathbf{e}, \neg(x\mathbf{u})). \tag{5}$$

Given the above observations, $MPE(\mathbf{e})$ will either be $MPE(\mathbf{e}, x\mathbf{u})$, $MPE(\mathbf{e}, \neg(x\mathbf{u}))$, or their union, depending on the value of parameter $\theta_{x|\mathbf{u}}$:

$$MPE(\mathbf{e}) = \begin{cases} MPE(\mathbf{e}, x\mathbf{u}), & \text{if } r(\mathbf{e}, x\mathbf{u}) \cdot \theta_{x|\mathbf{u}} > k(\mathbf{e}, x\mathbf{u}); \\ MPE(\mathbf{e}, \neg(x\mathbf{u})), & \text{if } r(\mathbf{e}, x\mathbf{u}) \cdot \theta_{x|\mathbf{u}} < k(\mathbf{e}, x\mathbf{u}); \\ MPE(\mathbf{e}, x\mathbf{u}) \cup MPE(\mathbf{e}, \neg(x\mathbf{u})), & \text{otherwise.} \end{cases}$$

Moreover, the MPE probability can always be expressed as:

$$MPE_p(\mathbf{e}) = \max(r(\mathbf{e}, x\mathbf{u}) \cdot \theta_{x|\mathbf{u}}, k(\mathbf{e}, x\mathbf{u})).$$

Figure 2 plots the relation between the MPE probability $MPE_p(\mathbf{e})$ and the value of parameter $\theta_{x|\mathbf{u}}$. According to the figure, if $\theta_{x|\mathbf{u}} > k(\mathbf{e}, x\mathbf{u})/r(\mathbf{e}, x\mathbf{u})$, i.e., region A of the plot, then we have $MPE(\mathbf{e}) = MPE(\mathbf{e}, x\mathbf{u})$, and thus the MPE solutions are consistent with $x\mathbf{u}$. Moreover, the MPE identity will remain unchanged as long as the value of $\theta_{x|\mathbf{u}}$ remains greater than $k(\mathbf{e}, x\mathbf{u})/r(\mathbf{e}, x\mathbf{u})$.

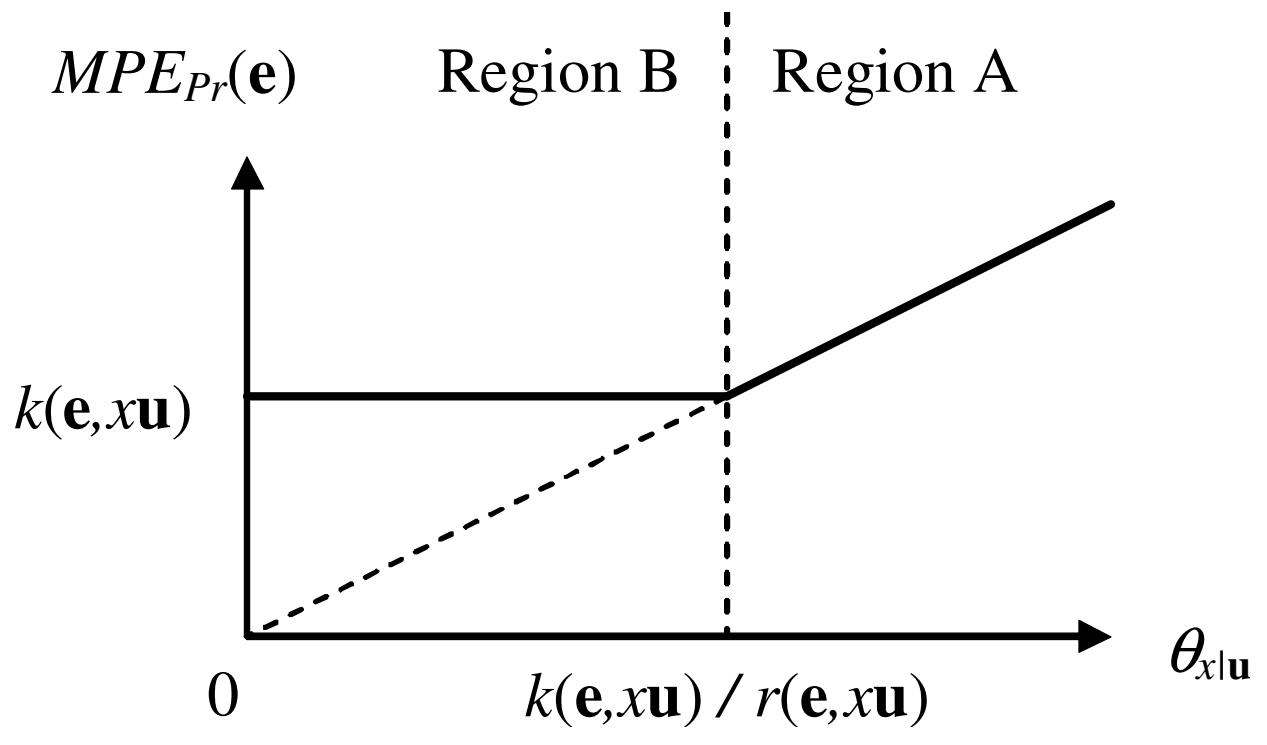

Figure 2: A plot of the relation between the MPE probability $MPE_p(\mathbf{e})$ and the value of parameter $\theta_{x|\mathbf{u}}$.

On the other hand, if $\theta_{x|\mathbf{u}} < k(\mathbf{e}, x\mathbf{u})/r(\mathbf{e}, x\mathbf{u})$, i.e., region B of the plot, then we have $MPE(\mathbf{e}) = MPE(\mathbf{e}, \neg(x\mathbf{u}))$, and thus the MPE solutions are inconsistent with $x\mathbf{u}$. Moreover, the MPE identity and probability will remain unchanged as long as the value of $\theta_{x|\mathbf{u}}$ remains less than $k(\mathbf{e}, x\mathbf{u})/r(\mathbf{e}, x\mathbf{u})$.

Therefore, $\theta_{x|\mathbf{u}} = k(\mathbf{e}, x\mathbf{u})/r(\mathbf{e}, x\mathbf{u})$ is the point where there is a change in the MPE identity if we were to change the value of parameter $\theta_{x|\mathbf{u}}$. At this point, $MPE(\mathbf{e}) = MPE(\mathbf{e}, x\mathbf{u}) \cup MPE(\mathbf{e}, \neg(x\mathbf{u}))$ and we have both MPE solutions consistent with $x\mathbf{u}$ and MPE solutions inconsistent with $x\mathbf{u}$. There are no other points where there is a change in the MPE identity. If we are able to find the constants $r(\mathbf{e}, x\mathbf{u})$ and $k(\mathbf{e}, x\mathbf{u})$ for the network parameter $\theta_{x|\mathbf{u}}$, we can then compute robustness conditions for MPE with respect to changes in this parameter.

# 4 Dealing with Co-Varying Parameters

The above analysis assumed that we can change a parameter $\theta_{x|\mathbf{u}}$ without needing to change any other parameters in the network. This is not realistic though in the context of Bayesian networks, where *co-varying parameters* need to add up to 1 for the network to induce a valid probability distribution. For example, if variable $X$ has two values, $x$ and $\bar{x}$, we must always have:

$$\theta_{x|\mathbf{u}} + \theta_{\bar{x}|\mathbf{u}} = 1.$$

We will therefore extend the analysis conducted in the previous section to account for the simultaneously changes in the co-varying parameters. We will restrict our attention to binary variables to simplify the discussion, but our results can be easily extended to multi-valued variables as we will show later.

In particular, assuming that we are changing parame-ters $\theta_{x|\mathbf{u}}$ and $\theta_{\bar{x}|\mathbf{u}}$ simultaneously for a binary variable $X$, we can now categorize all network instantiations which are consistent with evidence $\mathbf{e}$ into three groups, depending on whether they are consistent with $x\mathbf{u}$, $\bar{x}\mathbf{u}$, or $\neg\mathbf{u}$. Moreover, the most likely instantiations in each group are just $MPE(\mathbf{e}, x\mathbf{u})$, $MPE(\mathbf{e}, \bar{x}\mathbf{u})$, and $MPE(\mathbf{e}, \neg\mathbf{u})$ respectively. Therefore, if $\mathbf{x} \in MPE(\mathbf{e})$, then:

$$Pr(\mathbf{x}) = \begin{cases} r(\mathbf{e}, x\mathbf{u}) \cdot \theta_{x|\mathbf{u}}, & \text{if } \mathbf{x} \in MPE(\mathbf{e}, x\mathbf{u}); \\ r(\mathbf{e}, \bar{x}\mathbf{u}) \cdot \theta_{\bar{x}|\mathbf{u}}, & \text{if } \mathbf{x} \in MPE(\mathbf{e}, \bar{x}\mathbf{u}); \\ k(\mathbf{e}, \mathbf{u}), & \text{if } \mathbf{x} \in MPE(\mathbf{e}, \neg\mathbf{u}); \end{cases}$$

where:

$$\begin{aligned} r(\mathbf{e}, x\mathbf{u}) &= \frac{\partial MPE_p(\mathbf{e}, x\mathbf{u})}{\partial \theta_{x|\mathbf{u}}}; \\ r(\mathbf{e}, \bar{x}\mathbf{u}) &= \frac{\partial MPE_p(\mathbf{e}, \bar{x}\mathbf{u})}{\partial \theta_{\bar{x}|\mathbf{u}}}; \\ k(\mathbf{e}, \mathbf{u}) &= MPE_p(\mathbf{e}, \neg\mathbf{u}); \end{aligned}$$

and the MPE probability is:

$$MPE_p(\mathbf{e}) = \max(r(\mathbf{e}, x\mathbf{u}) \cdot \theta_{x|\mathbf{u}}, r(\mathbf{e}, \bar{x}\mathbf{u}) \cdot \theta_{\bar{x}|\mathbf{u}}, k(\mathbf{e}, \mathbf{u})).$$

Therefore, changing the co-varying parameters $\theta_{x|\mathbf{u}}$ and $\theta_{\bar{x}|\mathbf{u}}$ will not affect the identity of either $MPE(\mathbf{e}, x\mathbf{u})$ or $MPE(\mathbf{e}, \bar{x}\mathbf{u})$, nor will it affect the identity or probability of $MPE(\mathbf{e}, \neg\mathbf{u})$.

The robustness condition of an MPE solution can now be summarized as follows:

- If an MPE solution is consistent with $x\mathbf{u}$, it remains a solution as long as the following inequalities are true:

$$\begin{aligned} r(\mathbf{e}, x\mathbf{u}) \cdot \theta_{x|\mathbf{u}} &\geq r(\mathbf{e}, \bar{x}\mathbf{u}) \cdot \theta_{\bar{x}|\mathbf{u}}; \\ r(\mathbf{e}, x\mathbf{u}) \cdot \theta_{x|\mathbf{u}} &\geq k(\mathbf{e}, \mathbf{u}). \end{aligned}$$

- If an MPE solution is consistent with $\bar{x}\mathbf{u}$, it remains a solution as long as the following inequalities are true:

$$\begin{aligned} r(\mathbf{e}, \bar{x}\mathbf{u}) \cdot \theta_{\bar{x}|\mathbf{u}} &\geq r(\mathbf{e}, x\mathbf{u}) \cdot \theta_{x|\mathbf{u}}; \\ r(\mathbf{e}, \bar{x}\mathbf{u}) \cdot \theta_{\bar{x}|\mathbf{u}} &\geq k(\mathbf{e}, \mathbf{u}). \end{aligned}$$

- If an MPE solution is consistent with $\neg\mathbf{u}$, it remains a solution as long as the following inequalities are true:

$$\begin{aligned} k(\mathbf{e}, \mathbf{u}) &\geq r(\mathbf{e}, x\mathbf{u}) \cdot \theta_{x|\mathbf{u}}; \\ k(\mathbf{e}, \mathbf{u}) &\geq r(\mathbf{e}, \bar{x}\mathbf{u}) \cdot \theta_{\bar{x}|\mathbf{u}}. \end{aligned}$$

We note here that one can easily deduce whether an MPE solution is consistent with $x\mathbf{u}$, $\bar{x}\mathbf{u}$, or $\neg\mathbf{u}$ since it is a complete variable instantiation.

Therefore, all we need are the constants $r(\mathbf{e}, x\mathbf{u})$ and $k(\mathbf{e}, \mathbf{u})$ for each network parameter $\theta_{x|\mathbf{u}}$ in order to define robustness conditions for MPE. The constants $k(\mathbf{e}, \mathbf{u})$ can be easily computed from the constants $r(\mathbf{e}, x\mathbf{u})$ by observing the following:

$$\begin{aligned} k(\mathbf{e}, \mathbf{u}) &= MPE_p(\mathbf{e}, \neg\mathbf{u}) \\ &= \max_{\mathbf{u}^\star : \mathbf{u}^\star \neq \mathbf{u}} MPE_p(\mathbf{e}, \mathbf{u}^\star) \\ &= \max_{x\mathbf{u}^\star : \mathbf{u}^\star \neq \mathbf{u}} MPE_p(\mathbf{e}, x\mathbf{u}^\star) \\ &= \max_{x\mathbf{u}^\star : \mathbf{u}^\star \neq \mathbf{u}} r(\mathbf{e}, x\mathbf{u}^\star) \cdot \theta_{x|\mathbf{u}^\star}. \end{aligned} \tag{6}$$

As the algorithm we will describe later computes the $r(\mathbf{e}, x\mathbf{u})$ constants for all family instantiations $x\mathbf{u}$, the algorithm will then allow us to compute all the $k(\mathbf{e}, \mathbf{u})$ constants as well.

As a simple example, for the Bayesian network whose CPTs are shown in Figure 3, the current MPE solution without any evidence is $A = a, B = \bar{b}$, and has probability .4. For the parameters in the CPT of $B$, we can compute the corresponding $r(\mathbf{e}, x\mathbf{u})$ constants. In particular, we have $r(\mathbf{e}, ba) = r(\mathbf{e}, \bar{b}a) = r(\mathbf{e}, b\bar{a}) = r(\mathbf{e}, \bar{b}\bar{a}) = .5$ in this case. The $k(\mathbf{e}, \mathbf{u})$ constants can also be computed as $k(\mathbf{e}, a) = .3$ and $k(\mathbf{e}, \bar{a}) = .4$. Given these constants, we can easily compute the amount of change we can apply to co-varying parameters, say $\theta_{b|a}$ and $\theta_{\bar{b}|a}$, such that the MPE solution remains the same. The conditions we must satisfy are:

$$\begin{aligned} r(\mathbf{e}, \bar{b}a) \cdot \theta_{\bar{b}|a} &\geq r(\mathbf{e}, ba) \cdot \theta_{b|a}; \\ r(\mathbf{e}, \bar{b}a) \cdot \theta_{\bar{b}|a} &\geq k(\mathbf{e}, a). \end{aligned}$$

This leads to $\theta_{\bar{b}|a} \geq \theta_{b|a}$ and $\theta_{\bar{b}|a} \geq .6$. Therefore, the current MPE solution will remain so as long as $\theta_{\bar{b}|a} \geq .6$, which has a current value of .8.

We close this section by pointing out that our robustness equations can be extended to multi-valued variables as follows. If variable $X$ has values $x_1, \ldots, x_j$, with $j > 2$, then each of the conditions we showed earlier will consist of $j$ inequalities instead of just two. For example, if an MPE solution is consistent with $x_1\mathbf{u}$, it remains a solution as long as the following inequalities are true:

$$\begin{aligned} r(\mathbf{e}, x_1\mathbf{u}) \cdot \theta_{x_1|\mathbf{u}} &\geq r(\mathbf{e}, x^\star\mathbf{u}) \cdot \theta_{x^\star|\mathbf{u}} \text{ for all } x^\star \neq x_1; \\ r(\mathbf{e}, x_1\mathbf{u}) \cdot \theta_{x_1|\mathbf{u}} &\geq k(\mathbf{e}, \mathbf{u}). \end{aligned}$$

## 5 Computing Robustness Conditions

In this section, we will develop an algorithm for computing the constants $r(\mathbf{e}, x\mathbf{u})$ for all network parameters $\theta_{x|\mathbf{u}}$. In particular, we will show that they can be computed in time and space which is $O(n \exp(w))$, where $n$ is the number of network variables and $w$ is its treewidth.

| $A$ | $\Theta_A$ |
|---|---|
| $a$ | .5 |
| $\bar{a}$ | .5 |

| $A$ | $B$ | $\Theta_{B\|A}$ |
|---|---|---|
| $a$ | $b$ | .2 |
| $a$ | $\bar{b}$ | .8 |
| $\bar{a}$ | $b$ | .6 |
| $\bar{a}$ | $\bar{b}$ | .4 |

Figure 3: The CPTs for Bayesian network $A \longrightarrow B$.

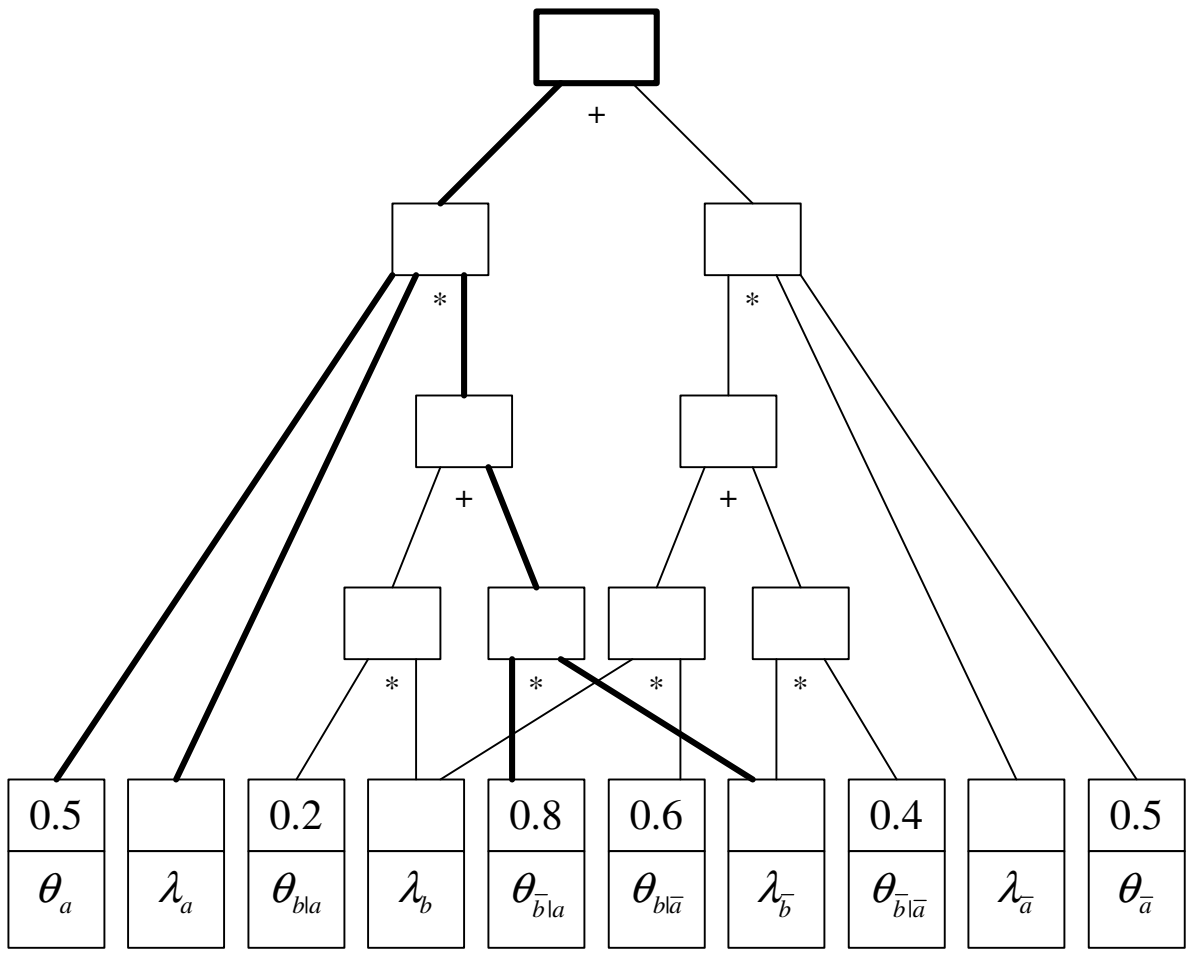

Figure 4: An arithmetic circuit for the above Bayesian network. The bold lines depict a complete sub-circuit, corresponding to the term $\lambda_a\lambda_{\bar{b}}\theta_a\theta_{\bar{b}|a}$.

### 5.1 Arithmetic Circuits

Our algorithm for computing the $r(\mathbf{e}, x\mathbf{u})$ constants is based on an *arithmetic circuit* representation of the Bayesian network [10]. Figure 4 depicts an arithmetic circuit for a small network consisting of two binary nodes, $A$ and $B$, shown in Figure 3. An arithmetic circuit is a rooted DAG, where each internal node corresponds to multiplication ($*$) or addition ($+$), and each leaf node corresponds either to a network parameter $\theta_{x|\mathbf{u}}$ or an evidence indicator $\lambda_x$; see Figure 4. Operationally, the circuit can be used to compute the probability of any evidence $\mathbf{e}$ by evaluating the circuit while setting the evidence indicator $\lambda_x$ to 0 if $x$ contradicts $\mathbf{e}$ and setting it to 1 otherwise. Semantically though, the arithmetic circuit is simply a factored representation of an exponential-size function that captures the network distribution. For example, the circuit in Figure 4 is simply a factored representation of the following function:

$$\lambda_a\lambda_b\theta_a\theta_{b|a} + \lambda_a\lambda_{\bar{b}}\theta_a\theta_{\bar{b}|a} + \lambda_{\bar{a}}\lambda_b\theta_{\bar{a}}\theta_{b|\bar{a}} + \lambda_{\bar{a}}\lambda_{\bar{b}}\theta_{\bar{a}}\theta_{\bar{b}|\bar{a}}.$$

This function, called the *network polynomial*, includes a *term* for each instantiation $\mathbf{x}$ of network variables, where the term is simply a product of the network parameters and evidence indicators which are consistent with $\mathbf{x}$. Moreover, the term for $\mathbf{x}$ evaluates to the probability value $Pr(\mathbf{e}, \mathbf{x})$ when the evidence indicators are set according to $\mathbf{e}$. Note that this function is multilinear. Therefore, a corresponding arithmetic circuit will have the property that two sub-circuits that feed into the same multiplication node will never contain a common variable. This property is important for some of the following developments.

### 5.2 Complete Sub-Circuits and Their Coefficients

Each term in the network polynomial corresponds to a *complete sub-circuit* in the arithmetic circuit. A complete sub-circuit can be constructed recursively from the root, by including all children of each multiplication node, and exactly one child of each addition node. The bold lines in Figure 4 depict a complete sub-circuit, corresponding to the term $\lambda_a\lambda_{\bar{b}}\theta_a\theta_{\bar{b}|a}$. In fact, it is easy to check that the circuit in Figure 4 has four complete sub-circuits, corresponding to the four terms in the network polynomial.

A key observation about complete sub-circuits is that if a network parameter is included in a complete sub-circuit, there is a unique path from the root to this parameter in this sub-circuit, even though there may be multiple paths from the root to this parameter in the original arithmetic circuit. This path is important as one can relate the value of the term corresponding to the sub-circuit and the parameter value by simply traversing the path as we show next.

Consider now a complete sub-circuit which includes a network parameter $\theta_{x|\mathbf{u}}$ and let $\alpha$ be the unique path in this sub-circuit connecting the root to parameter $\theta_{x|\mathbf{u}}$. We will now define the *sub-circuit coefficient* w.r.t. $\theta_{x|\mathbf{u}}$, denoted as $r$, in terms of the path $\alpha$ such that $r \cdot \theta_{x|\mathbf{u}}$ is just the value of the term corresponding to the sub-circuit.

Let $\Sigma$ be the set of all multiplication nodes on this path $\alpha$. The sub-circuit coefficient w.r.t. $\theta_{x|\mathbf{u}}$ is defined as the product of all children of nodes in $\Sigma$ which are themselves not on the path $\alpha$. Consider for example the complete sub-circuit highlighted in Figure 4 and the path from the root to the network parameter $\theta_a$. The coefficient w.r.t. $\theta_a$ is $r = \lambda_a\lambda_{\bar{b}}\theta_{\bar{b}|a}$. Moreover, $r \cdot \theta_a = \lambda_a\lambda_{\bar{b}}\theta_a\theta_{\bar{b}|a}$, which is the term corresponding to the sub-circuit.

### 5.3 Maximizer Circuits

An arithmetic circuit can be easily modified into a *maximizer circuit* to compute the MPE solutions, by simply replacing each addition node with a maximization node; see Figure 5. This corresponds to a circuit that computes the value of the maximum term in a network polynomial, instead of adding up the values of these terms. The value of the root will thus be the MPE probability $MPE_p(\mathbf{e})$. The maximizer circuit in Figure 5 is evaluated under evidence $A = a$, leading to an MPE probability of .4.

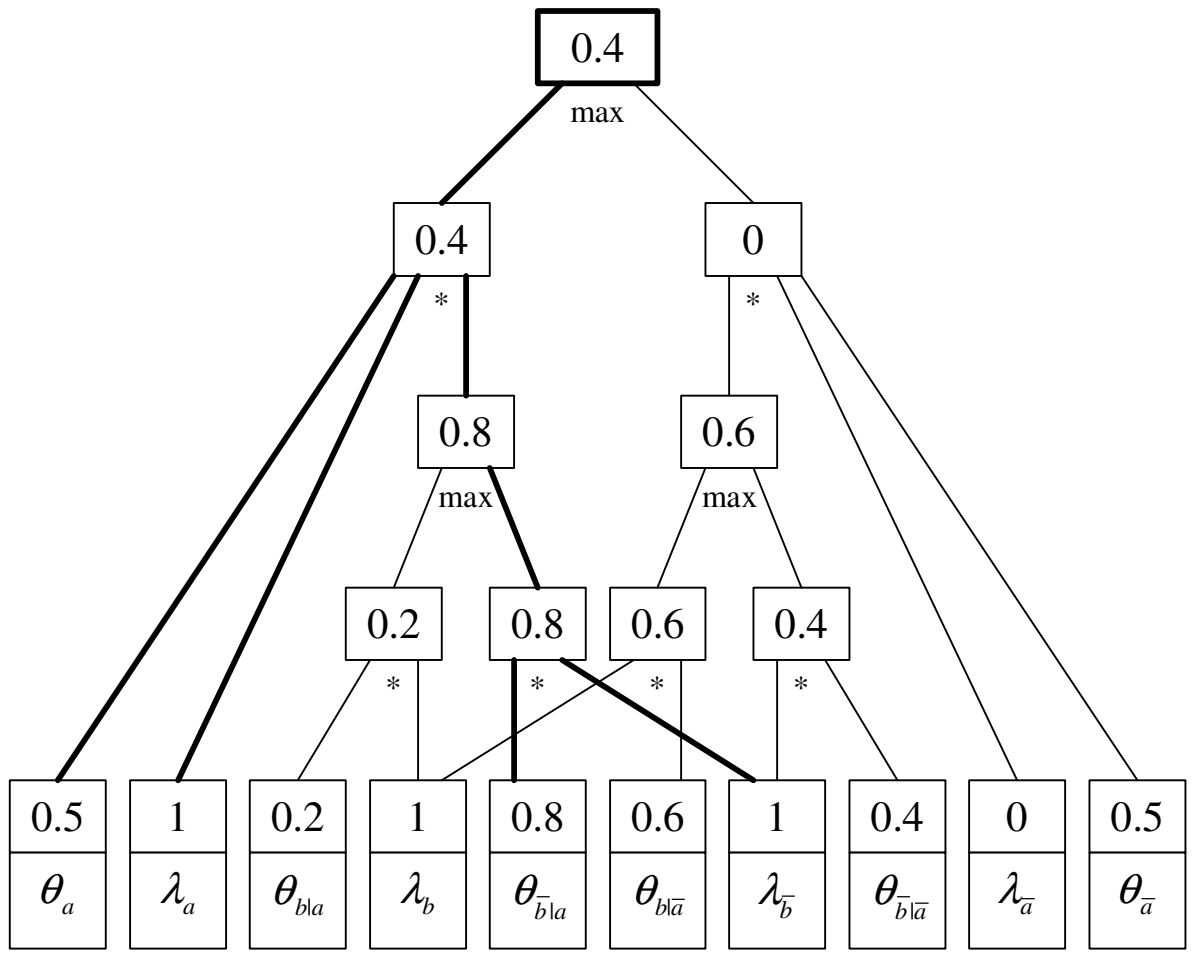

Figure 5: A maximizer circuit for a Bayesian network, evaluated under evidence $A = a$. Given this evidence, the evidence indicators are set to $\lambda_a = 1$, $\lambda_{\bar{a}} = 0$, $\lambda_b = 1$, $\lambda_{\bar{b}} = 1$. The bold lines depict the MPE sub-circuit.

To recover an MPE solution from a maximizer circuit, all we need to do is construct the *MPE sub-circuit* recursively from the root, by including all children of each multiplication node, and one child $c$ for each maximization node $v$, such that $v$ and $c$ have the same value; see Figure 5. The MPE sub-circuit will then correspond to an MPE solution. Moreover, if a parameter $\theta_{x|\mathbf{u}}$ is in the MPE sub-circuit, and the sub-circuit coefficient w.r.t $\theta_{x|\mathbf{u}}$ is $r$, then we have $r \cdot \theta_{x|\mathbf{u}}$ as the probability of MPE, $MPE_p(\mathbf{e})$.

Consider Figure 5 and the highlighted MPE sub-circuit, evaluated under evidence $A = a$. The term corresponding to this sub-circuit is $A = a, B = \bar{b}$, which is therefore an MPE solution. Moreover, we have two parameters in this sub-circuit, $\theta_a$ and $\theta_{\bar{b}|a}$, with coefficients $.8 = (1)(.8)$ and $.5 = (.5)(1)(1)$ respectively. Therefore, the MPE probability can be obtained by multiplying any of these coefficients with the corresponding parameter value, as $(.8)\theta_a = (.8)(.5) = .4$ and $(.5)\theta_{\bar{b}|a} = (.5)(.8) = .4$.

### 5.4 Computing $r(\mathbf{e}, x\mathbf{u})$

Suppose now that our goal is to compute $MPE(\mathbf{e}, x\mathbf{u})$ for some network parameter $\theta_{x|\mathbf{u}}$. Suppose further that $\alpha_1, \ldots, \alpha_m$ are all the complete sub-circuits that

**Algorithm 1** D-MAXC($\mathcal{M}$: a maximizer circuit, $\mathbf{e}$: evidence)

evaluate the circuit $\mathcal{M}$ under evidence $\mathbf{e}$; afterwards the value of each node $v$ is $p[v]$
$r[v] \leftarrow 1$ for root $v$ of circuit $\mathcal{M}$
$r[v] \leftarrow 0$ for all non-root nodes $v$ in circuit $\mathcal{M}$
**for** non-leaf nodes $v$ (parents before children) **do**
  **if** node $v$ is a maximization node **then**
    $r[c] \leftarrow \max(r[c], r[v])$ for each child $c$ of node $v$
  **if** node $v$ is a multiplication node **then**
    $r[c] \leftarrow \max(r[c], r[v] \prod_{c^\star} p[c^\star])$ for each child $c$ of node $v$, where $c^\star$ are the other children of node $v$

include $\theta_{x|\mathbf{u}}$. Moreover, let $\mathbf{x}_1, \ldots, \mathbf{x}_m$ be the instantiations corresponding to these sub-circuits and let $r_1, \ldots, r_m$ be their corresponding coefficients w.r.t. $\theta_{x|\mathbf{u}}$. It then follows that the probabilities of these instantiations are $r_1 \cdot \theta_{x|\mathbf{u}}, \ldots, r_m \cdot \theta_{x|\mathbf{u}}$ respectively. Moreover, it follows that:

$$MPE_p(\mathbf{e}, x\mathbf{u}) = \max_i r_1 \cdot \theta_{x|\mathbf{u}}, \ldots, r_m \cdot \theta_{x|\mathbf{u}},$$

and hence, from Equation 4:

$$\frac{\partial MPE_p(\mathbf{e}, x\mathbf{u})}{\partial \theta_{x|\mathbf{u}}} = r(\mathbf{e}, x\mathbf{u}) = \max_i r_1, \ldots, r_m.$$

Therefore, if we can compute the maximum of these coefficients, then we have computed the constant $r(\mathbf{e}, x\mathbf{u})$.

Algorithm 1 provides a procedure which evaluates the maximizer circuit and then traverses it top-down, parents before children, computing simultaneously the constants $r(\mathbf{e}, x\mathbf{u})$ for all network parameters. The procedure maintains an additional register value $r[.]$ for each node in the circuit, and updates these registers as it visits nodes. When the procedure terminates, it is guaranteed that the register value $r[\theta_{x|\mathbf{u}}]$ will be the constant $r(\mathbf{e}, x\mathbf{u})$. We will also see later that the register value $r[\lambda_x]$ is also a constant which provides valuable information for the MPE solutions. Figure 6 depicts an example of this procedure.

Algorithm 1 can be modelled as the all-pairs shortest path procedure, with edge $v \longrightarrow c$ having weight $0 = -\ln 1$ if $v$ is a maximization node, and weight $-\ln \pi$ if $v$ is a multiplication node, where $\pi$ is the product of the values of the other children $c^\star \neq c$ of node $v$. The length of the shortest path from the root to the network parameter $\theta_{x|\mathbf{u}}$ is then $-\ln r(\mathbf{e}, x\mathbf{u})$. It should be clear that the time and space complexity of the above algorithm is linear in the number of circuit nodes.[1] It is well known that we can compile a circuit for any Bayesian network in $O(n \exp(w))$ time and space, where $n$ is the number of network variables and $w$ is its treewidth [10]. Therefore, all constants $r(\mathbf{e}, x\mathbf{u})$ can be computed with the same complexity.

[1] More precisely, this algorithm is linear in the number of circuit nodes only if the number of children per multiplication node is bounded. If not, one can use a technique which gives a linear complexity by simply storing two additional bits with each multiplication node [10].

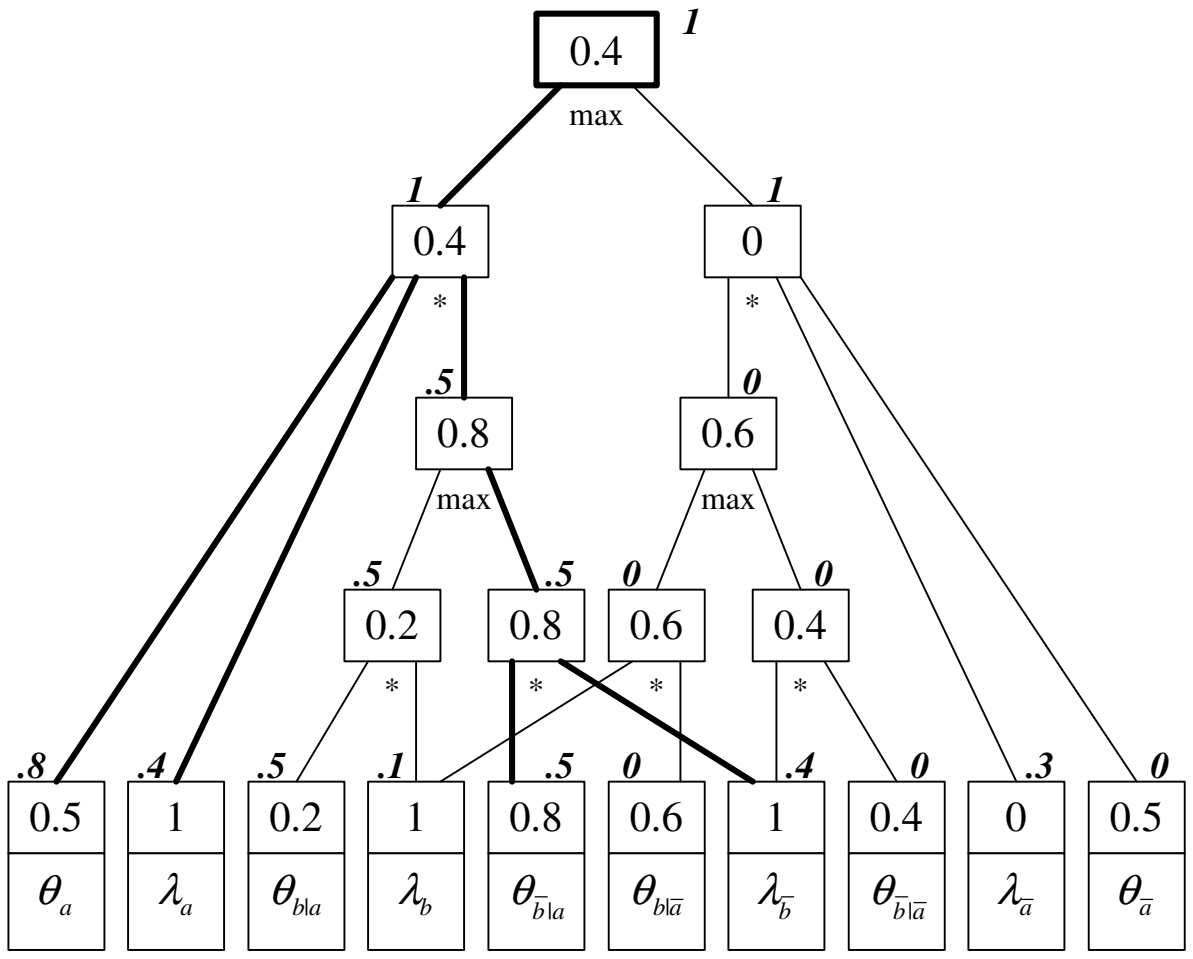

Figure 6: A maximizer circuit for a Bayesian network, evaluated under evidence $A = a$. Next to each node is the value $r[.]$ computed by Algorithm 1.

We close this section by pointing out that one can in principle use the jointree algorithm to compute $MPE_p(\mathbf{e}, x\mathbf{u}) = r(\mathbf{e}, x\mathbf{u}) \cdot \theta_{x|\mathbf{u}}$ for all family instantiations $x\mathbf{u}$ with the above complexity. In particular, by replacing summation with maximization in the jointree algorithm, one obtains $MPE_p(\mathbf{e}, \mathbf{c})$ for each cluster instantiation $\mathbf{c}$. Projecting on the families $X\mathbf{U}$ in cluster $\mathbf{C}$, one can then obtain $MPE_p(\mathbf{e}, x\mathbf{u})$ for all family instantiations $x\mathbf{u}$, which is all we need to compute robustness conditions for MPE.[2] Our method above, however, is more general for two reasons:

- The arithmetic circuit for a Bayesian network can be much smaller than the corresponding jointree by exploiting the local structures of the Bayesian network [12, 13].

- The constants computed by the algorithm for the evidence indicators can be used to answer additional MPE queries, which results after variations on the current evidence. This will be discussed in Section 7.

# 6 Example

We now go back to the example network in Figure 1, and compute robustness conditions for the current MPE solution using the inequalities we obtain in Section 4, and an implementation of Algorithm 1. After going through the CPT of each variable, our procedure found nine possible parameter changes that would produce a different MPE solution, as shown in Figure 7. From these nine suggested changes, only three changes make sense from a qualitative point of view:

- Decreasing the probability that the ignition is working from .9925 to at most .9133. (6th row)

- Decreasing the probability that the engine is working given both the battery and the ignition are working from .97 to at most .9108. (1st row)

- Decreasing the false-negative rate of the engine test from .09 to at most .0285. (9th row)

If we apply the first parameter change, we get a new MPE solution in which both the ignition and the engine are not working. If we apply either the second or third parameter change, we get a new MPE solution in which the engine is not working.

# 7 MPE under Evidence Change

We have discussed in Section 5.2 the notion of a complete sub-circuit and its coefficient with respect to a network parameter $\theta_{x|\mathbf{u}}$ which is included in the sub-circuit. In particular, we have shown how each sub-circuit corresponds to a term in the network polynomial, and that if a complete sub-circuit has coefficient $r$ with respect to parameter $\theta_{x|\mathbf{u}}$, then $r \cdot \theta_{x|\mathbf{u}}$ will be the value of the term corresponding to this sub-circuit.

The notion of a sub-circuit coefficient can be extended to evidence indicators as well. In particular, if a complete sub-circuit has coefficient $r$ with respect to an evidence indicator $\lambda_x$ which is included in the sub-circuit,

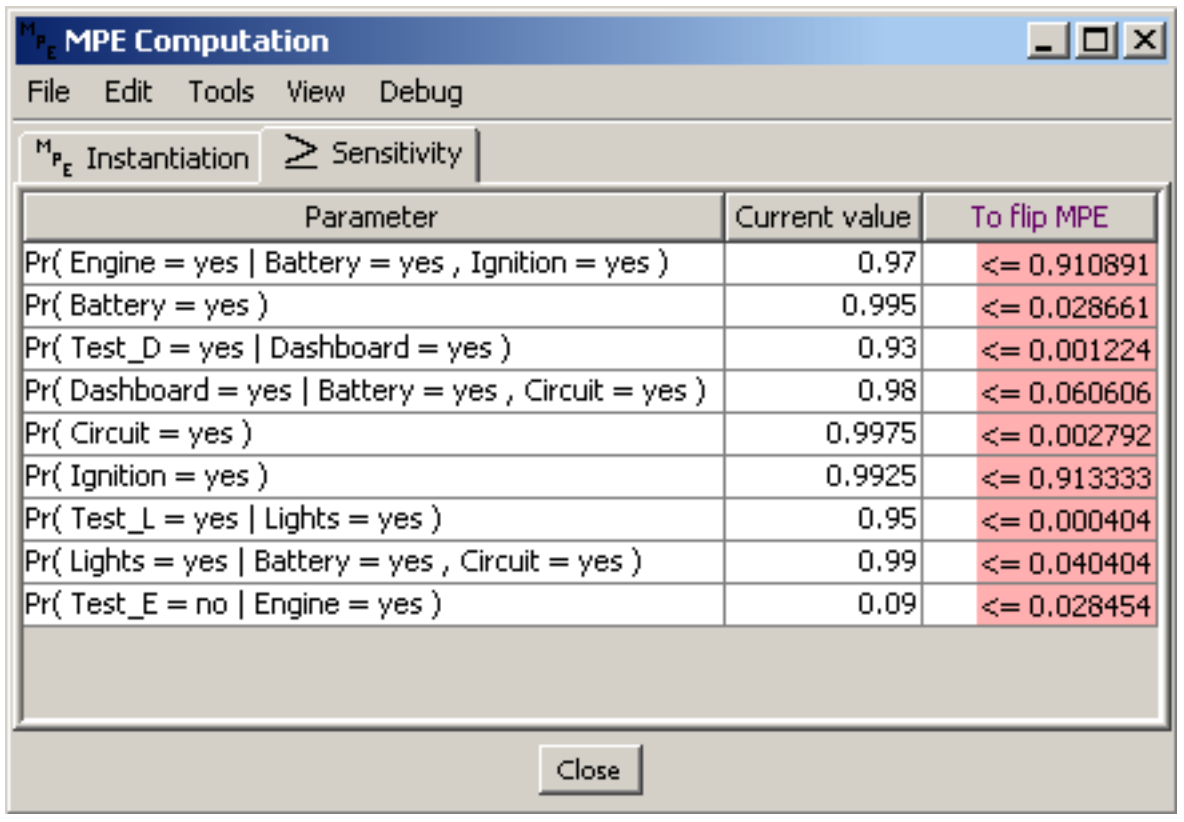

| Parameter | Current value | To flip MPE |
|---|---|---|
| Pr( Engine = yes \| Battery = yes , Ignition = yes ) | 0.97 | <= 0.910891 |
| Pr( Battery = yes ) | 0.995 | <= 0.028661 |
| Pr( Test_D = yes \| Dashboard = yes ) | 0.93 | <= 0.001224 |
| Pr( Dashboard = yes \| Battery = yes , Circuit = yes ) | 0.98 | <= 0.060606 |
| Pr( Circuit = yes ) | 0.9975 | <= 0.002792 |
| Pr( Ignition = yes ) | 0.9925 | <= 0.913333 |
| Pr( Test_L = yes \| Lights = yes ) | 0.95 | <= 0.000404 |
| Pr( Lights = yes \| Battery = yes , Circuit = yes ) | 0.99 | <= 0.040404 |
| Pr( Test_E = no \| Engine = yes ) | 0.09 | <= 0.028454 |

Figure 7: A list of parameter changes that would produce a different MPE solution.

[2] However, in case some of the parameters are equal to 0, one needs to use a special jointree [11].

then $r \cdot \lambda_x$ will be the value of the term corresponding to this sub-circuit.

Suppose now that $\alpha_1, \ldots, \alpha_m$ are all the complete sub-circuits that include $\lambda_x$. Moreover, let $\mathbf{x}_1, \ldots, \mathbf{x}_m$ be the terms corresponding to these sub-circuits and let $r_1, \ldots, r_m$ be their corresponding coefficients with respect to $\lambda_x$. It then follows that the values of these terms are $r_1 \cdot \lambda_x, \ldots, r_m \cdot \lambda_x$ respectively. Moreover, it follows that:

$$MPE_p(\mathbf{e} - X, x) = \max_i r_1, \ldots, r_m,$$

where $\mathbf{e} - X$ denotes the new evidence after having retracted the value of variable $X$ from $\mathbf{e}$ (if $X \in \mathbf{E}$, otherwise $\mathbf{e} - X = \mathbf{e}$). Therefore, if we can compute the maximum of these coefficients, then we have computed $MPE_p(\mathbf{e} - X, x)$. Note, however, that Algorithm 1 already computes the maximum of these coefficients for each $\lambda_x$ as the evidence indicators are nodes in the maximizer circuit as well, and therefore the register value $r[\lambda_x]$ gives us $MPE_p(\mathbf{e} - X, x)$ for every variable $X$ and value $x$.

Consider for example the circuit in Figure 6, and the coefficients computed by Algorithm 1 for the four evidence indicators. According to the above analysis, these coefficients have the following meanings:

| $\lambda_x$ | $\mathbf{e} - X, x$ | $r[\lambda_x] = MPE_p(\mathbf{e} - X, x)$ |
|---|---|---|
| $\lambda_a$ | $a$ | .4 |
| $\lambda_{\bar{a}}$ | $\bar{a}$ | .3 |
| $\lambda_b$ | $a, b$ | .1 |
| $\lambda_{\bar{b}}$ | $a, \bar{b}$ | .4 |

For example, the second row above tells us that the MPE probability would be .3 if the evidence was $A = \bar{a}$ instead of $A = a$. In general, if we have $n$ variables, we then have $O(n)$ variations on the current evidence of the form $\mathbf{e} - X, x$. The MPE probability of all of these variations are immediately available from the coefficients with respect to the evidence indicators.

The computation of these coefficients allows us to deduce the MPE identity after *evidence retraction*. In particular, suppose that variable $X$ is set as $x$ in evidence $\mathbf{e}$, and $MPE_p(\mathbf{e}) \geq MPE_p(\mathbf{e} - X, x^\star)$ for all other values $x^\star \neq x$. We can then conclude that $MPE_p(\mathbf{e}) = MPE_p(\mathbf{e} - X)$. Moreover, $MPE(\mathbf{e}) = MPE(\mathbf{e} - X)$ if $MPE_p(\mathbf{e}) > MPE_p(\mathbf{e} - X, x^\star)$ for all other values $x^\star \neq x$, or $MPE(\mathbf{e}) \subset MPE(\mathbf{e} - X)$ if there exists some $x^\star \neq x$ such that $MPE_p(\mathbf{e}) = MPE_p(\mathbf{e} - X, x^\star)$. Therefore, the current MPE solutions will remain so even after we retract $X = x$ from the evidence. This means that $X = x$ is not integral in the determination of the current MPE solutions given the other evidence, i.e., $\mathbf{e} - X$.

The result above also has implications on the identification of multiple MPE solutions given evidence $\mathbf{e}$. In particular, suppose that variable $X$ is not set in evidence $\mathbf{e}$, then:

- If the coefficients for the evidence indicators $\lambda_x$ and $\lambda_{\bar{x}}$ are equal, we must have both MPE solutions with $X = x$ and MPE solutions with $X = \bar{x}$. In fact, the coefficients must both equal the MPE probability $MPE_p(\mathbf{e})$ in this case.
- If the coefficient for the evidence indicator $\lambda_x$ is larger than the coefficient for the evidence indicator $\lambda_{\bar{x}}$, then every MPE solution must have $X = x$.

In the example above, we have $r[\lambda_{\bar{b}}] > r[\lambda_b]$, suggesting that every MPE solution must have $\bar{b}$ in this case.

# 8 Conclusion

We considered in this paper the problem of finding robustness conditions for MPE solutions of a Bayesian network under single parameter changes. We were able to solve this problem by identifying some interesting relationships between an MPE solution and the network parameters. In particular, we found that the robustness condition of an MPE solution under a single parameter change depends on two constants that are independent of the parameter value. We also proposed a method for computing such constants and, therefore, the robustness conditions of MPE in $O(n \exp(w))$ time and space, where $n$ is the number of network variables and $w$ is the network treewidth. Our algorithm is the first of its kind for ensuring the robustness of MPE solutions under parameter changes in a Bayesian network.

# Acknowledgments

This work has been partially supported by Air Force grant #FA9550-05-1-0075-P00002 and JPL/NASA grant #1272258. We would also like to thank James Park for reviewing this paper and making the observation on how to compute $k(\mathbf{e}, \mathbf{u})$ in Equation 6.